\documentclass[11pt]{article}

\usepackage[preprint]{acl}

\usepackage{times}
\usepackage{latexsym}

\usepackage[T1]{fontenc}

\usepackage[utf8]{inputenc}
\usepackage{hyperref}
\usepackage{url}
\usepackage{xcolor}
\usepackage{bm}
\usepackage{url}
\usepackage{booktabs}
\usepackage{bbding}
\usepackage{wrapfig}
\usepackage{multirow}
\usepackage{makecell}
\usepackage{pifont}
\usepackage{enumitem}
\usepackage{colortbl}
\usepackage{amsmath}
\usepackage{amssymb}
\usepackage{tikz}
\usepackage{array}
\usepackage[most]{tcolorbox}
\usepackage{subcaption}
\usepackage{minted}
\usepackage{lipsum}
\usepackage{tabularx}
\usepackage[skip=9pt]{caption}
\usepackage{pifont}

\usepackage{microtype}

\usepackage{inconsolata}
\usepackage{bbm}

\usepackage{graphicx}
\usepackage{arydshln}

\definecolor{darkblue}{RGB}{0, 51, 102}
\definecolor{lightbluebg}{RGB}{245, 248, 252}
\definecolor{blueframe}{RGB}{90, 150, 200}
\definecolor{dividercolor}{RGB}{200, 210, 220}
\definecolor{yellowtext}{RGB}{68,132,243}
\definecolor{yellowred}{RGB}{50,167,82}
\definecolor{yellowblue}{RGB}{251,191,5}
\definecolor{darkgreen}{rgb}{0,0.4,0}
\definecolor{maroon}{HTML}{A00000}
\definecolor{gray}{rgb}{0.5, 0.5, 0.5}
\definecolor{chocolate}{HTML}{D2691E}
\definecolor{indigo}{HTML}{4B0082}
\definecolor{violet}{HTML}{4B2E83}
\definecolor{lightgreen}{HTML}{E0FBE0}
\definecolor{lightred}{HTML}{FBE0E0}
\definecolor{cadmiumgreen}{rgb}{0.0, 0.42, 0.24}
\definecolor{forestgreen}{rgb}{0.13, 0.55, 0.13}
\definecolor{lightgray}{rgb}{0.9, 0.9, 0.9}
\definecolor{exp_table_blue}{HTML}{DAECED}
\definecolor{exp_table_yelow}{HTML}{F0B100}  
\definecolor{exp_table_red}{HTML}{CF292B} 
\definecolor{exp_table_green}{HTML}{05B04F}

\DeclareMathOperator*{\argmax}{arg\,max}

\title{Hint-Guided Diversified Policy Optimization for LLM Reasoning}

\author{
  \textbf{Zhiyu Cao\textsuperscript{1}\thanks{\ \ Work done during internship at Ant Group.}},
  \textbf{Kaixin Wu\textsuperscript{2}},
  \textbf{Mingjie Zhong\textsuperscript{2}},
  \textbf{Peifeng Li\textsuperscript{1}}\thanks{\ \ Corresponding author.},
  \textbf{Can Ye\textsuperscript{2}},
  \textbf{Qiaoming Zhu\textsuperscript{1}}
\\
  \textsuperscript{1}School of Computer Science and Technology, Soochow University, Suzhou, China\\
  \textsuperscript{2}Ant Group, Hangzhou, China\\
    \texttt{zycao18@stu.suda.edu.cn}, \texttt{\{pfli, qmzhu\}@suda.edu.cn}
}

\begin{document}
\maketitle
\begin{abstract}
Recent developments in Large Language Models (LLMs) have showcased impressive reasoning capabilities, with Reinforcement Learning with Verifiable Rewards (RLVR) being a promising enhancement strategy. However, existing reward mechanisms are constrained to the outcome-level correctness and lack explicit signals to guide the model to consider diverse solutions. In contrast, human problem solving typically involves evaluating multiple potential approaches and selecting the most reliable solution, a cognitive process that current RLVR frameworks do not explicitly incentivize. Inspired by this, we propose Hint-Guided Diversified Policy Optimization (HDPO), allowing the model to first list all potential candidate solution outlines as hints and then select the most reliable one for further reasoning. HDPO comprises two stages of \textbf{Cold Start for Structured Reasoning} and \textbf{Hint-Guided Diversified Reinforcement Learning} to incentivize the model to generate diverse and reliable solutions following the ``propose-select-think'' trajectory. Experimental results show that HDPO effectively boosts LLM reasoning and enhances the diversity of candidate solutions as well as the LLM's ability to identify reliable solutions.
\end{abstract}

\section{Introduction}
Recent research has yielded significant progress in enhancing the reasoning capabilities of Large Language Models (LLMs), exemplified by OpenAI-o1~\cite{Openai_o1}, DeepSeek-R1~\cite{Deepseek-r1}, and Kimi-1.5~\cite{Kimi_k1.5}, particularly for complex mathematical problem solving. A pivotal development underpinning these improvements is the adoption of Reinforcement Learning with Verifiable Reward (RLVR)~\cite{Dr.GRPO_Oat,LUFFY,Critique-grpo} as a core training paradigm. Within this framework, policy optimization algorithms are widely adopted to fine-tune LLMs, with Group Relative Policy Optimization (GRPO)~\cite{Deepseekmath} serving as a prominent instantiation of this approach.

\begin{figure}[t]
\begin{center}
 \includegraphics[width=\linewidth]{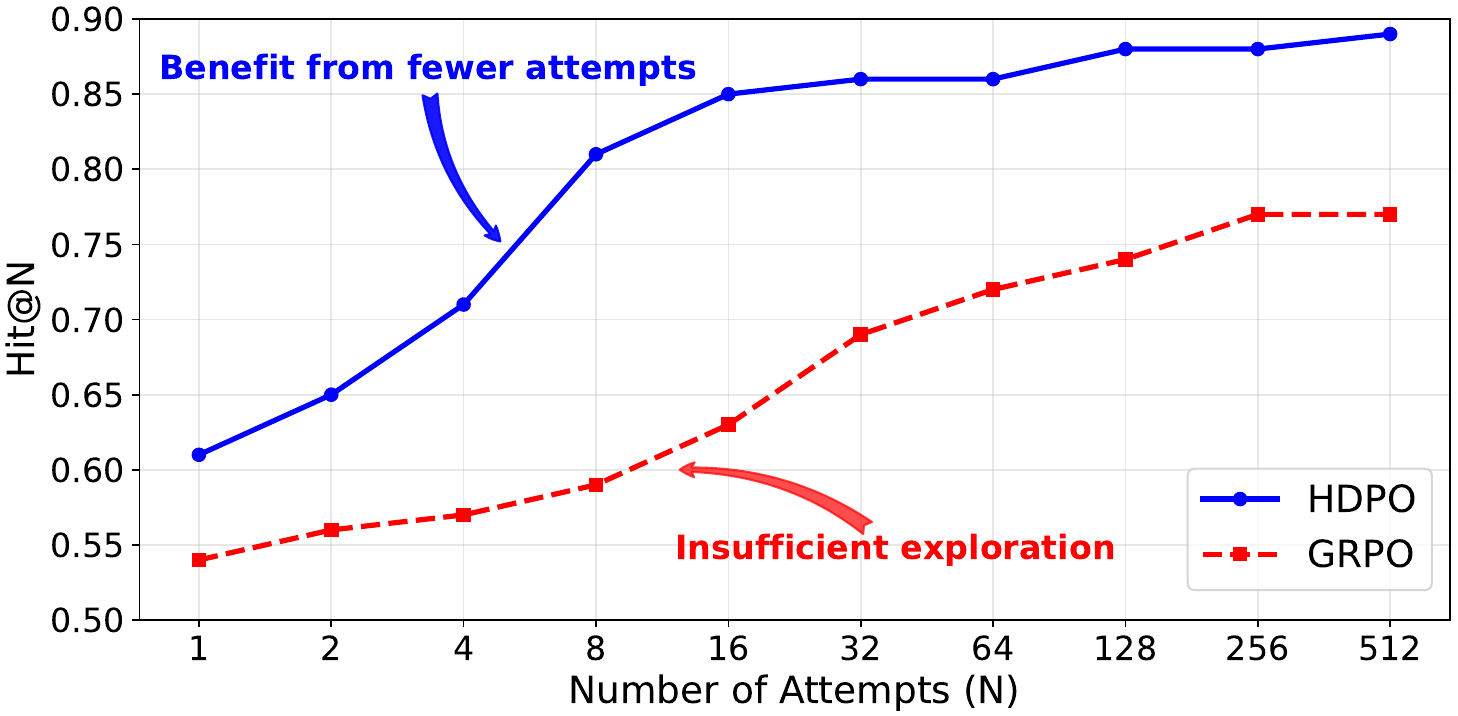}
 \caption{Hit rate of HDPO and GRPO on Olympiad-Bench under different number of attempts.}
 \label{fig:hit}
\end{center}
\vspace{-18pt}
\end{figure}

To ensure the credibility of complex reasoning, outcome-level rewards~\cite{gsm8k,Deepseekmath} are employed in RLVR to assess the correctness of the entire reasoning trajectory. However, this approach does not take into account the rationality of the process, and its sparse rewards may lead to homogenization of rewards, failing to provide comprehensive supervision. Therefore, recent studies~\cite{Edge-grpo,Critique-grpo,FCP} enhance the model's reasoning ability by providing more detailed feedback, such as feedback in natural language and process supervision.

\begin{figure*}[t]
\begin{center}
 \includegraphics[width=1\linewidth]{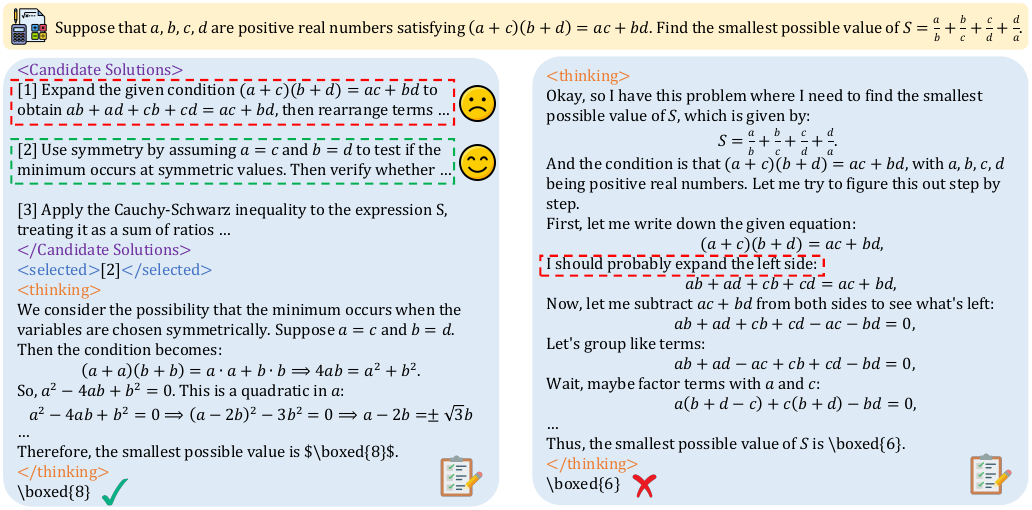}
 \caption{The comparison of HDPO's ``propose-select-think'' reasoning process (\textbf{left}) and the traditional reasoning process (\textbf{right}). More case studies can be found in Appendix~\ref{sec:case_study}.}
 \label{fig:case}
\end{center}
\vspace{-18pt}
\end{figure*}

We noticed that although the above methods have achieved notable success, they provide rewards based exclusively on correctness and thus fail to encourage the exploration of diverse solutions. Consequently, if the model gets trapped in a wrong solution, it may struggle to recover and identify the correct answer, reducing the fault tolerance of reasoning. As shown in Figure~\ref{fig:hit}, we evaluate the Hit@N accuracy of Qwen3-4B across varying numbers of sampling attempts. The results reveal that GRPO exhibits markedly lower performance when the number of attempts is limited, whereas HDPO consistently achieves superior accuracy even with fewer attempts. This suggests that the GRPO-trained model has insufficiently explored the space of diverse reasoning paths, which hinders its ability to generate correct answers within a limited number of trials. Ideally, the model should be able to explore a broader range of solutions and choose the most reliable solutions for further refinement. The key challenge lies in designing mechanisms that both incentivize the generation of diverse solutions and effectively assess their reliability.

To address this issue, we propose Hint-Guided Diversified Policy Optimization (HDPO), a novel approach to enhance the diversity and reliability of LLM reasoning. HDPO is built upon the ``propose-select-think'' reasoning trajectory, effectively stimulating the model to reason about diverse potential solutions. Specifically, our training methodology instills this capability in two stages: we first employ supervised fine-tuning (SFT) in the cold start stage to establish the model's proficiency in following the structured reasoning trajectory, followed by Reinforcement Learning (RL) to cultivate the policy model capable of generating diverse and reliable solutions. In the cold start phase, we employ an advanced LLM to construct such structured reasoning trajectories and apply filtering criteria based on correctness and reliability to ensure data quality. In the RL phase, we introduce diversity scheduling strategy and reliability reward to guide the model to actively explore all potential solutions.

As illustrated in Figure~\ref{fig:case}, the ``propose-select-think'' framework employed by HDPO offers a significant advantage over the traditional approach. While the conventional paradigm extends the given condition in a brute-force manner and arrives at an incorrect answer, HDPO promotes broader exploration of the solution space by generating multiple candidate solutions prior to engaging in detailed reasoning. This preliminary exploration enables the model to identify underlying symmetries, a critical insight that contributes to the correct answer. 

Crucially, HDPO advances existing paradigms through \textbf{policy internalization} and \textbf{joint optimization}. Unlike inference-time search methods~\cite{Self-Consistency,Tree_of_Thoughts} that incur high latency via repeated sampling, HDPO internalizes the explore-then-exploit mechanism into the policy parameters via RLVR, enabling zero-overhead, single-pass reasoning. Furthermore, departing from accuracy-centric RLVR, it directly mitigates solution homogenization and unreliable selection by jointly optimizing a scheduled diversity reward and a confidence-based reliability reward. To summarize, our contributions are as follows:
\begin{itemize}
\vspace{-3pt}
 \item We introduce HDPO to enhance LLM reasoning, a novel framework based on the ``propose-select-think'' structured reasoning process.
\vspace{-5pt}
\item The proposed diversity scheduling strategy and reliability reward enable LLMs to actively explore diverse and reliable solutions.
\vspace{-5pt}
\item Our extensive experimental results demonstrate that HDPO greatly boosts the model's reasoning ability.
\end{itemize}

\section{Related Work}
\paragraph{Chain-of-Thought for LLM Reasoning.}
Chain-of-thought~\cite{Chain-of-Thought} reasoning has shown promising progress in recent years. Building on this idea, tree-of-thought (ToT)~\cite{Tree_of_Thoughts} generalizes reasoning into structured sequences of thoughts and determines the next action through self-evaluation.~\citet{Least_to_Most_Prompting} proposes least-to-most prompting by decomposing problems, enabling language models to solve problems that are more difficult than those in the prompts. Graph of Thoughts (GoT)~\cite{Graph_of_Thoughts} models LLM reasoning as an arbitrary graph, where thoughts are represented as vertices and dependencies between thoughts are represented as edges. Unlike ToT/GoT, which require repeated sampling and heuristic scoring at inference, HDPO internalizes the propose-select-think paradigm into the policy weights via reward-shaped RL, yielding zero-overhead inference. Through structured reinforcement learning, the model learns to generate diverse candidate hints and autonomously select the most reliable trajectory in a single forward pass.

\paragraph{Reinforcement Learning for LLM Reasoning.}
Recent advancements in reinforcement learning (RL)~\cite{Deepseekmath,Vapo,AlphaMed} have made significant progress in enhancing complex reasoning, shifting from surface-level response generation to sophisticated problem-solving. ~\citet{InstructGPT} utilize preference data to train reward models and apply Proximal Policy Optimization (PPO)~\cite{PPO} to perform reinforcement learning on LLMs for alignment with human preferences. Subsequently, numerous methods have employed reinforcement learning to align LLMs with human preferences~\cite{d-RLAIF,MDO,Text2Reward}. DeepSeekMath~\cite{Deepseekmath} proposes the GRPO algorithm, which simplifies the training process of reinforcement learning and significantly enhances the performance of LLMs in the mathematical domain through RL. Building on this foundation, Dr.GRPO~\cite{Dr.GRPO_Oat} proposes an unbiased optimization method that improves token efficiency while maintaining reasoning performance. ~\citet{L1} use length constraints to ensure that the reasoning language model adheres to the specified length by the user. To ensure that sequence-level rewarding and optimization are aligned, Group Sequence Policy Optimization (GSPO)~\cite{GSPO} defines importance ratios based on sequence likelihood.

\begin{figure*}[t]
\begin{center}
 \includegraphics[width=1\linewidth]{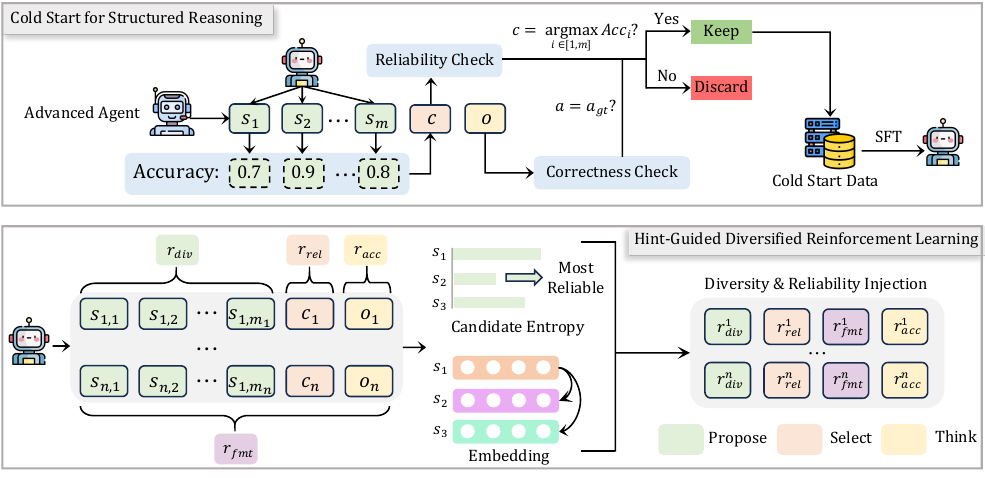}
 \vspace{-18pt}
 \caption{Overview of Hint-Guided Diversified Policy Optimization, comprising two stages: (1) \textbf{Cold Start for Structured Reasoning}, which constructs correctness- and reliability-filtered ``propose-select-think'' trajectories via an advanced LLM; and (2) \textbf{Hint-Guided Diversified Reinforcement Learning}, which optimizes the policy via diversity scheduling and reliability rewards to foster solution exploration.}
 \label{fig:method}
\end{center}
\vspace{-14pt}
\end{figure*}

\section{Methods}
\subsection{Preliminary}
We utilize GRPO (Shao et al. 2024) to optimize the policy model $\pi_\theta$. Given a prompt $p$, the old policy $\pi_{\theta_{\mathrm{old}}}$ generates $G$ responses $\{x_1,\ldots,x_G\}$ with rewards $\{r_1,\ldots,r_G\}$. The response-level advantages are computed through group normalization:
\begin{equation}
\hat{A}_i=
\frac{
r_i-\operatorname{mean}(r_1,\ldots,r_G)
}{
\operatorname{std}(r_1,\ldots,r_G)+\epsilon_{\mathrm{norm}}
}, 
\label{eq:advantage}
\end{equation}
where $\epsilon_{\mathrm{norm}}$ is a small constant for numerical stability.

The policy is then optimized using the token-level clipped GRPO objective with KL regularization:
{\small \begin{align}
\mathcal{L}_{\mathrm{GRPO}}(\theta)
={}&
-\frac{1}{G}
\sum_{i=1}^{G}
\frac{1}{T_i}
\sum_{t=1}^{T_i}
\Bigg[
\min\Big(
\rho_{i,t}(\theta)\hat{A}_i,
\notag\\
\operatorname{clip}\big(
\rho_{i,t}(\theta),1-&\epsilon,1+\epsilon
\big)\hat{A}_i
\Big)
-\beta D_{\mathrm{KL}}\Big(
\pi_\theta
\,\big\|\,
\pi_{\mathrm{ref}}
\Big)
\Bigg], 
\label{eq:loss_grpo}
\end{align}}
where $T_i$ is the length of response $x_i$, and $\rho_{i,t}(\theta)=\frac{\pi_\theta(x_{i,t}\mid p,x_{i,<t})}{\pi_{\theta_{\mathrm{old}}}(x_{i,t}\mid p,x_{i,<t})}$ is the token-level importance ratio. All tokens in $x_i$ share the response-level advantage $\hat A_i$. The hyperparameters $\epsilon$ and $\beta$ control clipping and KL regularization, respectively. GRPO improves LLM reasoning by rewarding positive relative advantages while constraining policy drift.

\subsection{Overview}
As shown in Figure~\ref{fig:method}, HDPO establishes a two-stage optimization policy. To enable the model to first propose candidate solutions and then select the most reliable solution for thinking, we perform \textbf{Cold Start for Structured Reasoning} in the first stage. Specifically, we distill the data of the ``propose-select-think'' reasoning trajectory from an advanced LLM to perform SFT. To ensure the validity of these trajectories, we filter the distilled data based on both correctness and reliability. In the second stage, we propose \textbf{Hint-Guided Diversified Reinforcement Learning} to enhance the quality of model reasoning through the diversity scheduling strategy and the reliability reward during the process of policy optimization.

\subsection{Cold Start for Structured Reasoning}
\label{cold_start}
Previous reasoning LLMs~\cite{Dr.GRPO_Oat,LUFFY,Critique-grpo} typically involve generating a single solution directly from a given problem, a paradigm that lacks an explicit mechanism to encourage exploration of the diverse solution space. Therefore, we seek to teach the model to follow the ``propose-select-think'' reasoning trajectory. Specifically, the model initially generates an overview comprising multiple candidate solutions, each delineated in several sentences, and then selects the most reliable candidate to serve as a guiding hint for subsequent thinking.

To fulfill the need for the policy model to follow the reasoning pattern of ``propose-select-think'', we distill such trajectory data from an advanced teacher LLM. Given a problem $q$, the teacher model is prompted to generate a structured reasoning trajectory that adheres to the ``propose-select-think'' framework. The prompt for the structured reasoning can be found in Appendix~\ref{appendix:prompt_reasoning}. Each trajectory is of the form $\tau=\{s_1, s_2, ..., s_m, c, o\}$, where $s_i$, $c$, and $o$ respectively represent the $i$-th proposed candidate solution, the most reliable solution selected, and the thinking process. To ensure the rationality of the generated data, we apply a two-fold filtering criterion: (1) the correctness of the final answer and (2) the reliability of the selected solution.

We extract the final answer $a$ from the thinking process $o$, and compare it with the ground truth $a_{gt}$. If there is an error, we discard this piece of data. Otherwise, we proceed to check the reliability of the selected solution. Specifically, we feed the problem and the $m$ candidate solution overviews sequentially into a lightweight model $\pi_s$, allowing it to sample $L$ answers for each candidate solution and calculate the accuracy $Acc_i$ of the sampled answers $\mathcal{A}_i$ corresponding to each candidate solution:
\begin{align}
\mathcal{A}_i&=\{\hat{a}_{i,j}\}_{j=1}^L \sim \pi_s(q,s_i), \\
Acc_i&=\frac{\sum_{\hat{a}_{i,j}\ \in \mathcal{A}_i} \mathbbm{1}[\hat{a}_{i,j}=a_{gt}]}{|\mathcal{A}_i|}.
\end{align}
Our key insight is that more reliable candidate solutions are also conducive to the model generating accurate answers. If the optimal candidate solution $c$ selected by the model aligns with the candidate solution exhibiting the highest accuracy, i.e., $c=\argmax_{i \in [1,m]} Acc_i$, then it indicates that the solution selected by the model is reliable, and this piece of data is retained.

After constructing the cold start training data $\mathcal{D}_{cssr} = \{(q_i, \tau_i)\}_{i=1}^n$, we optimize the policy model $\pi_\theta$ through supervised fine-tuning:
\begin{equation}
\small
\hat{\theta} = \arg\min_{\theta} -\mathbb{E}_{(q,\tau)\sim\mathcal D_{\mathrm{cssr}}} \left[ \log \pi_\theta(\tau\mid q) \right], 
\end{equation}
where $\hat{\pi}_{\theta}$ is the policy model after supervised fine-tuning during the cold start phase.

\subsection{Hint-Guided Diversified Reinforcement Learning}
\label{HDRL}
Following the cold-start phase, the policy model has acquired the structured reasoning ability of ``propose-select-think''. To further enhance its ability to navigate a broader solution space and select high-quality candidates, we refine the hints associated with candidate solutions through reinforcement learning. For each problem $q$, the model $\hat{\pi}_{\theta}$ samples $G$ reasoning trajectories in each rollout. The form of the $i$-th reasoning trajectory is $x_i = \{s_{i,1},s_{i,2},...,s_{i,m_i},c_i,o_i\}$, where $m_i$ represents the number of candidate solutions generated in the $i$-th reasoning trajectory. For notational convenience, we henceforth omit the subscript corresponding to each trajectory, i.e., each reasoning trajectory is denoted as $x=\{s_{1},s_{2},...,s_{m},c,o\}$.

\noindent \textbf{Format Reward.} To ensure that the model's output follows the predefined ``propose-select-think'' structure, we define the format reward. We determine (1) whether the model's output contains the tags ``\texttt{<Candidate Solutions></Candidate Solutions>}'', ``\texttt{<selected></selected>}'', and ``\texttt{<thinking></thinking>}''; (2) whether the number of candidate solution overview is between 1 and 5; (3) whether exactly one reliable solution is selected. A reward of $r_{fmt} = 0.1$ is granted for strict adherence to the structure, while any deviation results in a reward of $r_{fmt} = 0$.

\noindent \textbf{Accuracy Reward.} To ensure the accuracy of the final answer, we extract the answer $a$ from the model's reasoning trace $o$, and judge whether it is consistent with the ground truth $a_{gt}$:
\vspace{-1pt}
\begin{equation}
r_{acc} = 
\begin{cases} 
1, & \text{if } a = a_{gt} , \\
0, & \text{if } a \neq a_{gt}.
\end{cases}
\end{equation}
Therefore, the model receives positive feedback when its output matches the ground truth.

\noindent \textbf{Diversity Reward.} We aim to encourage the model to explore a broader solution space rather than being confined to a single solution, as this facilitates the selection of more reliable solutions. Therefore, the candidate solutions should be pairwise distinct to ensure the diversity of solutions. To quantify the similarity among candidate solutions, candidate solutions are encoded into embeddings $\mathcal{E} = \{e_1, \dots, e_m\}$, and similarity is quantified as the average pairwise cosine similarities:
\vspace{-1pt}
\begin{equation}
\small
\alpha_{sim} = \frac{1}{m (m-1)}\sum_{i \in [1,m]} \sum_{j \in [1,m] \textbackslash \{i\}} \cos (e_i,e_j), 
\end{equation}
which is defined for $m\ge2$. For $m=1$, pairwise diversity is unobserved, so we use the mean $\bar d=|\mathcal V|^{-1}\sum_{j\in\mathcal V}(1-\alpha_{sim}^{(j)})$ of correct multi-candidate rollouts $\mathcal V$ in the same group as a neutral reference. If $\mathcal V$ is empty, the group receives no diversity term. In the early stages of training, the policy model should prioritize solution correctness to establish a solid foundation of reasoning. Once its ability is sufficiently strong, we gradually shift the optimization objective toward promoting solution diversity. To this end, we have designed a diversity scheduling strategy, using sine learning coefficients to gradually increase the proportion of diversity rewards during the training process. Therefore, the diversity reward for the $t$-th step in the training process is as follows: 
\vspace{-1pt}
\begin{align}
\small
r_{div}^{(i)} = &\mu\,d_i\sin\left(\frac{\pi}{2}\cdot\frac{\max(0,t-t_{wp})}{t_{max}-t_{wp}}\right), \\
 &d_i=\begin{cases}1-\alpha_{sim}^{(i)},&m_i\ge2,\\\bar d,&m_i=1.\end{cases}
\end{align}
where $\mu$ is the hyperparameter controlling the level of diversity, $t_{wp}$ is the number of warm-up steps before the diversity reward is applied and $t_{max}$ is the maximum number of training steps. Under this scheduling strategy, the weight coefficient of diversity gradually increases from $0$ to $\mu$. As the model's capability improves, we refine its reasoning process to generate diverse candidate solutions while preserving answer correctness.

\begin{table*}[t!]
\centering
\vspace{-0.1in}
\resizebox{\textwidth}{!}{
\begin{tabular}{lcccccccccc}
\toprule
    \textbf{Methods} &   \textbf{AMC 23} & \textbf{AIME 24} & \textbf{AIME 25} & \textbf{Math} & {\textbf{Minerva}} & \textbf{Olympiad} & {\textbf{GPQA}} & \textbf{MMLU-Pro} & {\textbf{SciBench}} & \textbf{Avg.} \\
\midrule
\multicolumn{9}{@{}l}{\textit{Previous RLVR Methods}} \\
\quad EDGE-GRPO    &  73.79 & 16.36 & 19.37 & 84.10 & 49.89 & 48.79 & 46.76 & 54.10	& 25.79  &  46.55  \\
\quad Oat    &  73.59 & 38.91 & 16.64 & 85.34 & 38.75 & 50.82 & 44.43  & 47.72	& 26.31  &  46.95 \\
\quad SimpleRL    & 74.27  & 36.43 & 16.34 & 85.22 & 40.43 & 49.84 &  47.31  & 46.48	& 25.72 &  46.89\\
\quad LUFFY    &  79.68 & 28.95 & 26.61 & 87.99 & 44.54 & 56.09 & 61.51 & 67.63	& 33.37  & 54.04   \\
\quad Critique-GRPO    &  81.88 & 44.93 & 28.44 & 91.22 & 51.51 & 61.46 & 67.28  & 74.31 &	41.38 &  60.27  \\
\midrule
\multicolumn{9}{@{}l}{\textit{Qwen3-4B}} \\
\quad Base Model    &  59.10 & 20.36 & 18.73 & 78.53 & 39.82 & 40.11 & 46.81  &  57.95  &  21.93   &  42.59  \\
\quad GRPO & 73.32 & 32.08 & 25.78 & 87.01 & 49.17 & 55.50 & 58.47  &  68.57  &  33.92   &  53.76    \\
\quad \cellcolor{lightgreen}HDPO & \cellcolor{lightgreen}\textbf{83.26} & \cellcolor{lightgreen}\textbf{43.56} & \cellcolor{lightgreen}\textbf{27.92} & \cellcolor{lightgreen}\textbf{92.47} & \cellcolor{lightgreen}\textbf{54.92} & \cellcolor{lightgreen}\textbf{62.37} & \cellcolor{lightgreen}\textbf{65.49}  & \cellcolor{lightgreen}\textbf{76.12}  &  \cellcolor{lightgreen}\textbf{43.16}   &   \cellcolor{lightgreen}\textbf{61.03}   \\
\midrule
\multicolumn{9}{@{}l}{\textit{Qwen2.5-Math-7B}} \\
\quad Base Model     & 71.08 & 31.55 & 15.04 & 79.49 & 30.17 & 43.73 & 48.72  &  36.74 & 17.70  & 41.58   \\
\quad GRPO  & 80.92 & 40.76 & 24.55 & 89.72 & 44.86 & 53.68 & 60.18  & 53.19	 & 35.54 &  53.71  \\
\quad \cellcolor{lightgreen}HDPO & \cellcolor{lightgreen}\textbf{85.79} & \cellcolor{lightgreen}\textbf{48.75} & \cellcolor{lightgreen}\textbf{31.69} & \cellcolor{lightgreen}\textbf{93.28} & \cellcolor{lightgreen}\textbf{53.54} & \cellcolor{lightgreen}\textbf{64.76} & \cellcolor{lightgreen}\textbf{69.89}  & \cellcolor{lightgreen}\textbf{65.63} & \cellcolor{lightgreen}\textbf{44.36}  & \cellcolor{lightgreen}\textbf{61.97}   \\
\midrule
\multicolumn{9}{@{}l}{\textit{DS-R1-Llama-8B}} \\
\quad Base Model     & 63.78 & 16.43 & 16.08 & 77.54 & 37.67 & 38.88 & 41.29  &  48.57  &  19.03   &  39.92  \\
\quad GRPO & 76.62 & 24.55 & 22.32 & 85.53 & 45.14 & 54.74 & 58.60    &   59.53  &  36.10   &   51.46   \\
\quad \cellcolor{lightgreen}HDPO & \cellcolor{lightgreen}\textbf{84.17} & \cellcolor{lightgreen}\textbf{37.76} & \cellcolor{lightgreen}\textbf{30.75} & \cellcolor{lightgreen}\textbf{92.74} & \cellcolor{lightgreen}\textbf{56.68} & \cellcolor{lightgreen}\textbf{56.19} & \cellcolor{lightgreen}\textbf{63.41}  & \cellcolor{lightgreen}\textbf{67.90}  & \cellcolor{lightgreen}\textbf{41.02}  & \cellcolor{lightgreen}\textbf{58.96}   \\
\bottomrule
\end{tabular}}
\caption{Performance of HDPO and baselines on nine reasoning benchmarks.}
\label{tab:main_results}
\vspace{-10pt}
\end{table*}

\noindent \textbf{Reliability Reward.} During the reinforcement learning phase, it is essential to maintain diversity among candidate solutions while simultaneously ensuring that the solution selected by the model is the most reliable. However, assessing reliability via the LLM-based sampling method described in Section~\ref{cold_start} would incur substantial computational overhead. Therefore, we employ the entropy of the candidate solutions generated by the model as a proxy to judge the reliability of the solutions. Candidate solutions with lower entropy reflect higher model confidence and are more likely to lead to the correct answer. The uncertainty of each candidate solution is quantified by computing the mean entropy across its constituent tokens: 
\begin{align}
H_t &= -\sum_{v=1}^{|V|} p_v^{(t)} \log p_v^{(t)}, 
&
\bar{H} &= \frac{1}{T} \sum_{t=1}^{T} H_t, 
\end{align}
where $p_v^{(t)}$ is the probability of the $v$-th token in the vocabulary $V$ at the $t$-th position, $H_t$ is the entropy corresponding to the $t$-th token in the candidate solution, and $T$ is the number of tokens in the candidate solution. The candidate solutions $s_1$ to $s_m$ are sorted in ascending order of entropy to obtain $\mathcal{S}=\{\hat{s}_1, \hat{s}_2,...,\hat{s}_m\}$, then the reliability reward corresponding to this reasoning trajectory is $r_{rel}$:
\begin{equation}
r_{rel} = \frac{1}{k}, \ \ \mathrm{s.t.} \  s_c=\hat{s}_k, 
\end{equation}
which means that if the most reliable solution $c$ chosen by the model is ranked $k$-th among all candidate solutions, the reliability reward is $r_{rel} = \frac{1}{k}$. In this manner, the model is incentivized to select candidate solutions that exhibit greater reliability.

By incorporating the aforementioned four rewards, we guide the model to adhere to the ``propose-select-think'' trajectory, generating diverse and reliable solutions. The overall reward of the $i$-th sample in the rollout is as follows,
\begin{equation}
\small
r_i = r_{fmt}^{(i)} + r_{acc}^{(i)} + \mathbbm{1}[r_{acc}^{(i)}>0] \cdot r_{div}^{(i)} + \bar{r}_{\text{acc}} \cdot r_{rel}^{(i)}, 
\end{equation}
where $\mathbbm{1}[r_{acc}^{(i)}>0]$ indicates whether the answer corresponding to the $i$-th sample is correct, and HDPO applies the diversity reward exclusively when the answer generated by the model is correct. To calibrate the reliability reward and mitigate overconfidence, we scale $r_{\text{rel}}^{(i)}$ by the group-level average accuracy $\bar{r}_{\text{acc}} = \frac{1}{G}\sum_{j=1}^{G} r_{\text{acc}}^{(j)}$. Subsequently, the group-relative advantage is computed based on Equation~\ref{eq:advantage}, and the model $\hat{\pi}_{\theta}$ is optimized using the clipped loss function specified in Equation~\ref{eq:loss_grpo}.

\section{Experiments}
\subsection{Experimental Setup}
\noindent \textbf{Datasets and Benchmarks.} In the cold-start phase, DAPO-Math-17k~\cite{Dapo} was utilized as seed data to generate thinking trajectories. Subsequently, DeepScaleR-Preview-Dataset~\cite{Deepscaler} was employed for the reinforcement learning stage. We evaluated HDPO on nine reasoning benchmarks: AMC 2023~\cite{amc}, AIME 2024~\cite{aime24}, AIME 2025~\cite{aime25}, Math-500~\cite{math}, Minerva~\cite{Minerva}, OlympiadBench~\cite{OlympiadBench}, MMLU-Pro~\cite{MMLU-Pro}, SciBench~\cite{SciBench}, and GPQA-Diamond~\cite{gpqa}.

\noindent \textbf{Implementation Details.} We experimented with using \texttt{Qwen3-4B}~\cite{Qwen3}, \texttt{Qwen2.5-Math-7B}~\cite{Qwen2.5-math}, and \texttt{DeepSeek-R1-Distill-Llama-8B}~\cite{Deepseek-r1} as backbone models. In order to construct cold start data based on seed data, we employ \texttt{Qwen3-235B-A22B-Instruct-2507} to generate thought trajectories and another lightweight model \texttt{Qwen3-1.7B} to measure the reliability of each candidate solution. More implementation details can be found in Appendix~\ref{appendix:implementation}.

\noindent \textbf{Baselines and Evaluation.} We compared HDPO with the standard GRPO trained on DAPO-Math-17k and DeepScaleR-Preview-Dataset. Additionally, we also use the following powerful RLVR-based baselines: Qwen2.5-Math-7B-Oat-Zero~\cite{Dr.GRPO_Oat}, Qwen-2.5-7B-SimpleRL-Zoo~\cite{Simplerl-zoo}, LUFFY-Qwen-Math-7B-Zero~\cite{LUFFY}, EDGE-GRPO-Qwen-7B~\cite{Edge-grpo} and 
Critique-GRPO-8B~\cite{Critique-grpo}. For each question in the test set, we sample 16 responses and calculate the unbiased pass@3~\cite{unbiased_pass} metric for evaluation.

\subsection{Main Results}
We conducted evaluations on different model scales (4B, 7B, and 8B) and model families (Qwen and Llama). As shown in Table~\ref{tab:main_results}, HDPO outperforms the RLVR methods on both mathematical reasoning and general reasoning benchmarks. It suggests that HDPO can be adapted to different LLMs. Additionally, since GRPO was employed for model optimization, we also compared with the original GRPO method. It can be observed that HDPO has a significant advantage over GRPO, with improvements of 7.27 and 8.26 respectively on Qwen3-4B and Qwen2.5-Math-7B, which is due to HDPO's ability to encourage the model to explore more potential solutions, thereby choosing more reliable solutions and leading to accurate answers.

It is worth noting that, after optimization with HDPO, Qwen3-4B outperforms Critique-GRPO built on the larger Qwen3-8B backbone, demonstrating the effectiveness of HDPO beyond model scaling. Moreover, HDPO consistently improves not only mathematical reasoning but also general reasoning performance, suggesting strong robustness and generalization across different reasoning domains.

\subsection{Ablation Study}
Taking \texttt{Qwen2.5-Math-7B} as an example, we analyze each module of HDPO in Table~\ref{tab:ablation_results}.

\noindent \textbf{Cold Start and Reinforcement Learning.} We experimented with removing the cold start phase (``w/o Cold Start'') and directly performing reinforcement learning. The results indicate that the cold start effectively enhances the model's reasoning ability, attributed to its adaptation of the model to the ``propose-select-think'' reasoning pattern. Additionally, we used a relatively small LLM \texttt{Qwen3-14B} to construct the cold start data (``CS-Qwen3-14B''). Although performance decreased compared to using a stronger model, it still surpassed that of GRPO, indicating that higher-quality cold-start data generated by more capable models leads to better initialization. We also analyzed the scenario of directly removing the entire reinforcement learning phase (``w/o RL'') and only performing cold start. The performance dropped compared to removing cold start phase, highlighting the importance of diversity and reliability.

\begin{table}[t!]
\centering
\resizebox{\linewidth}{!}{
\begin{tabular}{lcccc}
\toprule
    \textbf{Methods} &  \textbf{AIME 25} & \textbf{Math} & \textbf{Olympiad} & {\textbf{GPQA}} \\
\midrule

HDPO & \textbf{31.69} & \textbf{93.28} & \textbf{64.76} & \textbf{69.89}  \\

~~w/o Cold Start & 28.63 & 90.55 & 61.59 & 65.74  \\

~~w/o RL & 21.47 & 85.81 & 51.23 & 58.13  \\

~~w/o Diversity Reward & 27.24 & 90.20 & 60.16 & 66.29    \\

~~w/o Diversity Scheduling & 30.51 & 91.77 & 63.43 & 68.17  \\

~~w/o Reliability Reward & 28.86 & 90.38 & 61.46 & 66.51 \\

\midrule
CS-Qwen3-14B & 29.76 & 91.24 & 62.41 & 67.70 \\
Random Reliability & 27.45 & 89.58 & 60.73 & 65.47 \\
Reverse Reliability & 26.93 & 88.60 & 60.14 & 63.82 \\
0/1 Reward & 29.87 & 91.30 & 63.15 & 67.19 \\
\bottomrule
\end{tabular}
}
\caption{Ablation results for each component in HDPO.}
\label{tab:ablation_results}
\vspace{-9.4pt}
\end{table}

\noindent \textbf{Diversity.} During the reasoning process, we encourage the model to actively explore diverse solution spaces by imposing the diversity reward. As shown in Table~\ref{tab:ablation_results}, the model performance decreases significantly after removing the diversity reward (``w/o Diversity Reward''), underscoring the importance of promoting diversity among candidate solutions to foster more thorough reasoning and improve answer accuracy. Recognizing that the model exhibits limited reasoning capability in the early stages of RL training, HDPO gradually increases the proportion of diversity rewards through a diversity scheduling strategy. We can observe that its contribution to HDPO is positive, indicating that such a scheduling strategy is beneficial for balancing the accuracy and diversity.

\noindent \textbf{Reliability.} 
To unlock the model's ability to select reliable solutions, we designed the reliability reward that checks whether the solution selected by the model is optimal. After removing the reliability reward, the model's performance declined because the model lacked the supervisory signal for the selection step and was unable to choose effective candidate solution.

Reliability reward in HDPO is built on the foundation that the reliability and confidence of a solution are closely related. We also considered two other variants of the reliability reward. The first alternative randomly assigns the reliability reward of a sample to one of the values in $[1, \frac{1}{2}, ..., \frac{1}{m}]$ (i.e., ``Random Reliability''), where $m$ is the number of candidate solutions. The other method sorts the confidence of the candidate solutions in descending order (i.e., ``Reverse Reliability''), and the reliability reward is the reciprocal of the rank of the solution selected by the model. Both variants show inferior performance compared to the reliability reward design in HDPO, verifying that solutions with high confidence are favorable for reasoning.

In order to take into account the different degrees of reliability of the candidate solutions chosen by the model, we designed continuous rewards, which are inversely proportional to the ranking of the selected solutions. An alternative approach assigns a binary reward of 1 if the model selects the top-ranked solution and 0 otherwise. As shown by ``0/1 Reward'', this discrete reward scheme degrades performance, which stems from its failure to discriminate between solutions of differing reliability.

\subsection{Analysis}
\noindent \textbf{Trends in Accuracy and Diversity.} 
To maintain the diversity of candidate solutions generated by the model, we introduced diversity reward to encourage differences between solutions. We compared the trend of diversity of candidate solutions and the accuracy of answers during the training process before and after adding diversity rewards.

\begin{figure}[t]
\begin{center}
 \includegraphics[width=1\linewidth]{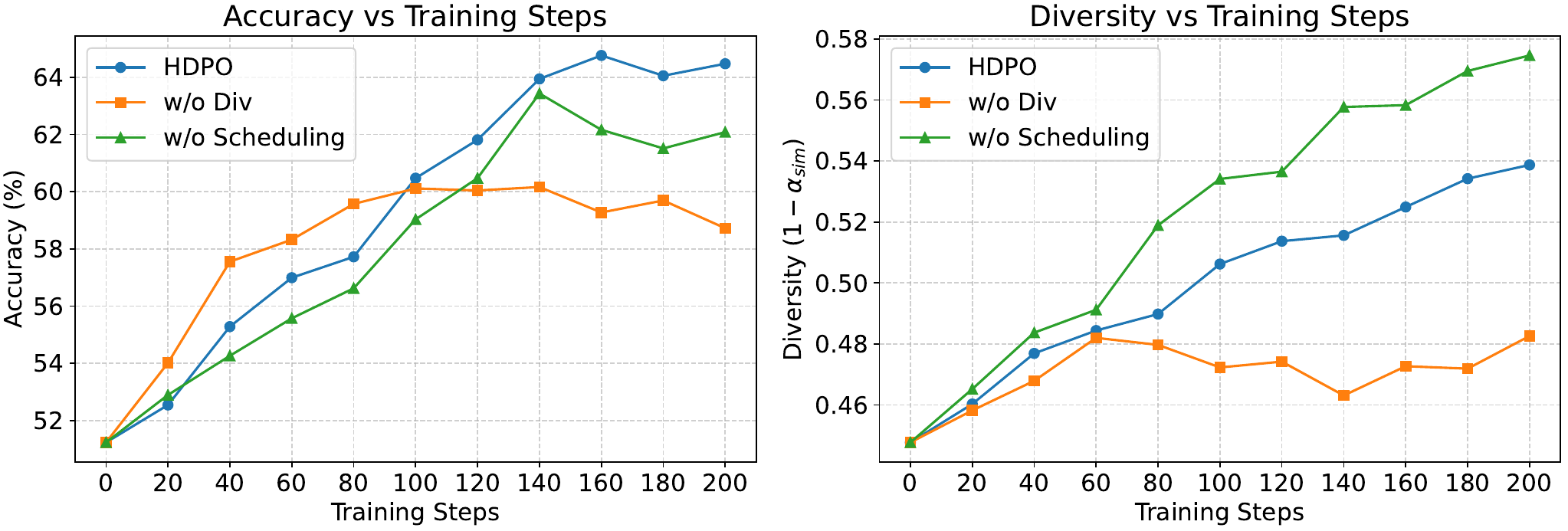}
 \caption{The effect of diversity reward configurations on model performance for Olympiad-Bench.}
 \label{fig:div}
\end{center}
\vspace{-12pt}
\end{figure}

As shown in Figure~\ref{fig:div}, compared to the HDPO with diversity reward (i.e., ``HDPO''), the performance of HDPO without adding diversity reward (i.e., ``w/o Div'') reached saturation prematurely. This phenomenon results from the diversity reward directing the model toward a broader solution space rather than constraining optimization to outcome-level correctness. Furthermore, we examined the effect of ablating the diversity scheduling strategy (``w/o Scheduling''). The absence of scheduling mechanism compromises training stability, which in turn causes the model to overemphasize diversity. Although accuracy initially peaks, it subsequently exhibits fluctuations, ultimately yielding inferior performance compared to the scheduled variant.

\begin{table}[t!]
\centering
\resizebox{\linewidth}{!}{
\begin{tabular}{lccc}
\toprule
    \textbf{Types} &   \textbf{Qwen3-4B} & \textbf{Qwen2.5-Math-7B} & \textbf{DS-R1-Llama-8B} \\
\midrule
Spearman    &  0.64 & 0.67 & 0.62  \\
Kendall    & 0.61  & 0.63 & 0.57  \\
\bottomrule
\end{tabular}}
\caption{Correlation between confidence and reliability.}
\label{tab:correlation}
\vspace{-0.3cm}
\end{table}

\begin{table}[t]
    \centering

\resizebox{\linewidth}{!}{
\begin{tabular}{ccccc}
\specialrule{1pt}{1pt}{1pt}
\multicolumn{1}{c}{\textbf{Backbones}}                                 & \textbf{AIME 25} & \textbf{Math} & \textbf{Olympiad} & \textbf{GPQA}      \\ \midrule
\multicolumn{5}{c}{\cellcolor[HTML]{9dbcd4}\textit{\textbf{w/o Reliability Reward}}}                                                                        \\ \midrule
\multicolumn{1}{c}{Qwen3-4B}    & 60.63             & 71.19             & 64.79          & 66.19            \\
\multicolumn{1}{c}{Qwen2.5-Math-7B} & 61.46             & 67.74             & 62.25          & 72.13            \\
\multicolumn{1}{c}{DS-R1-Llama-8B} & 57.71             & 66.54             & 58.54          & 63.04            \\ \midrule
\multicolumn{5}{c}{\cellcolor[HTML]{05ffa6}\textit{\textbf{w/ Reliability Reward}}}                                                                                 \\ \midrule
\multicolumn{1}{c}{Qwen3-4B}    & 75.64             & 87.40             & 80.58          & 81.50            \\
\multicolumn{1}{c}{Qwen2.5-Math-7B}  & 77.29    & 91.65    & 82.27 & 82.17   \\ 
\multicolumn{1}{c}{DS-R1-Llama-8B}  & 71.88    & 82.51    & 76.87 & 75.32   \\ 
\bottomrule[1pt]
\end{tabular}
}
    \caption{Assessing the impact of the reliability reward on candidate selection accuracy.}
    \label{table:change_reliability_reward}
    \vspace{-0.3cm}
\end{table}

\noindent \textbf{Correlation between Solution Confidence and Reliability.} 
HDPO treats solution confidence as a proxy for reliability, assuming a strong correlation between the two. To validate this, we sampled 10,000 training rollouts and ranked candidates using both confidence and the LLM-sampled accuracy metric (Section~\ref{cold_start}). As shown in Table~\ref{tab:correlation}, confidence and reliability demonstrate a strong positive correlation, with correlation coefficients exceeding 0.5 across all three backbones.

We further assess the impact of the reliability reward on selection behavior. Specifically, we extracted the candidate solutions chosen by the model and measured their reliability using the method described in Section~\ref{cold_start} to assess whether the model's selection aligns with the most reliable solution. As shown in Table~\ref{table:change_reliability_reward}, after applying the reliability reward, the model is able to select reliable solutions more accurately, aligning with the reliability derived from accuracy rate. This indicates that the reliability reward effectively stimulates the model's capacity to identify reliable candidate solutions.

\begin{table}[t!]
\centering
\resizebox{\linewidth}{!}{
\begin{tabular}{lcccc}
\toprule
    \textbf{Methods} &   \textbf{AIME 25} & \textbf{Math} & \textbf{Olympiad} & \textbf{GPQA} \\
\midrule
HDPO (Iter 1)    &  29.32 & 90.15 & 61.98 & 67.04  \\
HDPO (Iter 2)    & 30.46 & 90.77 & 62.84 & 68.15   \\
HDPO (Iter 3)    &  31.63 & 91.75 & 63.96 & 69.23  \\
HDPO (Iter 4)    &  31.32 & 92.33 & 63.85 & 69.06   \\
HDPO (Iter 5)    &  31.43 & 92.14 & 63.89 & 69.48   \\
\bottomrule
\end{tabular}}
\caption{Experimental results from the first five iterations of the self-evolution process.}
\label{tab:self_evolution}
\vspace{-0.6cm}
\end{table}

\noindent \textbf{Self-Evolution Alternating between Cold Start and Reinforcement Learning.} 
In the cold start phase, we employed an advanced LLM to construct structured reasoning trajectories following the ``propose-select-think'' framework. To alleviate the dependence on the teacher model, we designed a self-evolving policy optimization scheme. Specifically, during the cold start phase, the policy model $\pi_\theta$ serves as its own teacher model. In each iteration, we first employ $\pi_\theta$ to curate cold start data and perform supervised fine-tuning on itself, and then proceed with reinforcement learning. This process is repeated over multiple sequential iterations.

As presented in Table~\ref{tab:self_evolution}, we observed that after five iterations, the model's performance is almost comparable to that of using \texttt{Qwen3-235B-A22B-Instruct-2507} as the teacher model to construct cold-start data. This indicates that our HDPO can also stimulate the reasoning ability of LLMs through a self-evolving iterative loop, thereby offering a scalable approach to continuously optimize the capability of the policy model.

Additional analyses are provided in the appendices, covering diversity metrics and trade-offs (Appendices~\ref{appendix:div_measure} and~\ref{appendix:balance_div_rel}), model variants and alternative optimization algorithms (Appendices~\ref{appendix:base_vs_ins} and~\ref{appendix:adapt_policy_alg}), hyperparameter and latency studies (Appendices~\ref{appendix:maximum_number_of_candidates} and~\ref{appendix:infer_latency}), as well as baseline comparisons and the positioning of HDPO (Appendices~\ref{appendix:comparison_other_paradigms} and~\ref{appendix:positioning_of_hdpo}).

\section{Conclusion}
In this paper, we propose Hint-Guided Diversified Policy Optimization (HDPO), explicitly guiding the model to expand the exploration space of reasoning. HDPO leverages the ``propose-select-think'' reasoning paradigm, wherein multiple candidate solutions are generated as hints, and the most promising candidate is selected for thinking. Experimental results on multiple benchmarks show that HDPO greatly enhances the reasoning ability of LLMs.

\section*{Limitations}
Although HDPO can promote the diversity of solutions generated by the model in policy optimization, there are still two limitations. First, we inject diversity in the form of hints, but there may be some incorrect solutions among the candidate solutions generated by the model, which could affect the subsequent reasoning of the model. How to improve the model's robustness to candidate solutions is a direction worth exploring. On the other hand, HDPO employs the confidence of candidate solutions to estimate reliability. How to design other estimation strategies and integrate the results of multiple strategies to enhance the estimation of reliability is left for future work.

\section*{Acknowledgements}
This research was supported by the National Natural Science Foundation of China (Nos. 62276177 and 62376181), and Project Funded by the Priority Academic Program Development of Jiangsu Higher Education Institutions. This work was also supported by Ant Group Research Intern Program.

\bibliography{custom}

\newpage

\appendix

\section{Prompt for Structured Reasoning}
\label{appendix:prompt_reasoning}
We provide the prompt used for data construction in the cold start phase in Figure~\ref{fig:prompt_reasoning}. We default to allowing the teacher model to generate up to $\mathcal{M} = 5$ candidate solutions. Each solution overview includes 1 to 5 sentences, containing high-level solution strategies and concepts.

\begin{table}[t!]
\centering
\vspace{-0.1in}
\resizebox{\linewidth}{!}{
\begin{tabular}{lcccc}
\toprule
    \textbf{Methods} &   \textbf{AIME 25} & \textbf{Math} & \textbf{Olympiad} & \textbf{GPQA} \\
\midrule
HDPO (0.6B)    &  31.69 & 93.28 & 64.76 & 69.89  \\
HDPO (4B)    & 32.12 & 93.85 & 65.54 & 70.26   \\
HDPO (bge-m3)    &  31.24 & 92.73 & 64.48 & 69.31  \\
HDPO (B\&R)    &  29.76 & 92.01 & 62.81 & 67.76   \\
HDPO (LLM-Div)	&  32.51	&  92.14	&  64.22	&  70.47  \\
\bottomrule
\end{tabular}}
\caption{Comparison of different diversity measurement methods.}
\label{tab:div_measurement}
\end{table}

\section{Implementation Details}
\label{appendix:implementation}
For the data construction in the cold start phase, we sample $L = 10$ answers each time to evaluate reliability. After filtering, 83,279 data entries were retained for supervised fine-tuning. During the reinforcement learning phase, we sample $G = 8$ responses in each rollout, with the maximum length of the prompt set to 1024 and the maximum length of the response set to 3072. To calculate the diversity reward, we employ \texttt{Qwen3-Embedding-0.6B}~\cite{Qwen3_Embedding} as $\pi_s$ to encode each candidate solution and set $\mu=0.5$ to control the degree of diversity. The maximum training steps $t_{max}$ is set to 200, and the number of warm-up steps $t_{wp}$ before applying the diversity reward is set to 20. All our experiments were conducted on a server equipped with 8 $\times$ NVIDIA A100 (80GB) GPUs.

\section{Solution Diversity Measurement}
\label{appendix:div_measure}
To improve computational efficiency, HDPO employs \texttt{Qwen3-Embedding-0.6B} to encode candidate solutions and calculates cosine similarity to measure the diversity of solutions. To further investigate alternative approaches, we also analyzed encoding with \texttt{Qwen3-Embedding-4B} and \texttt{bge-m3}~\cite{Bge_m3_embedding}, as well as measuring similarity by calculating the sum of pairwise BLEU$_1$ and ROUGE$_l$ scores (i.e., HDPO (B\&R)). As shown in Table~\ref{tab:div_measurement}, the more powerful the embedding model, the more beneficial it is for model training, attributed to the fact that the embedding model directly influences the feedback for the diversity of candidate solutions generated by the model. Additionally, the method based on lexical matching does not take into account semantic-level information, demonstrate significantly poorer performance in diversity measurement compared to embedding-based approaches.

Another optional approach is to prompt the LLM to determine the similarity between candidate solutions. Specifically, we prompt \texttt{Qwen3-4B-Instruct-2507} to determine the similarity between two candidate solutions. The prompt we use is shown in Figure~\ref{fig:LLM_Div}. Next, we parse out the probability that the model outputs ``Yes'' as a measure of the similarity between the two candidate solutions. As shown by ``HDPO (LLM-Div)'' in Table~\ref{tab:div_measurement}, the performance of this approach is comparable to that of using the embedding representation to measure the semantic similarity. This indicates that semantic similarity can effectively reflect the correlation between solutions. However, the method of using the LLM to judge the relevance of solutions will lead to more computational overhead, so we adopt the approach of calculating the semantic similarity based on the embedding representation.

\begin{figure}[t]
\begin{center}
 \includegraphics[width=1\linewidth]{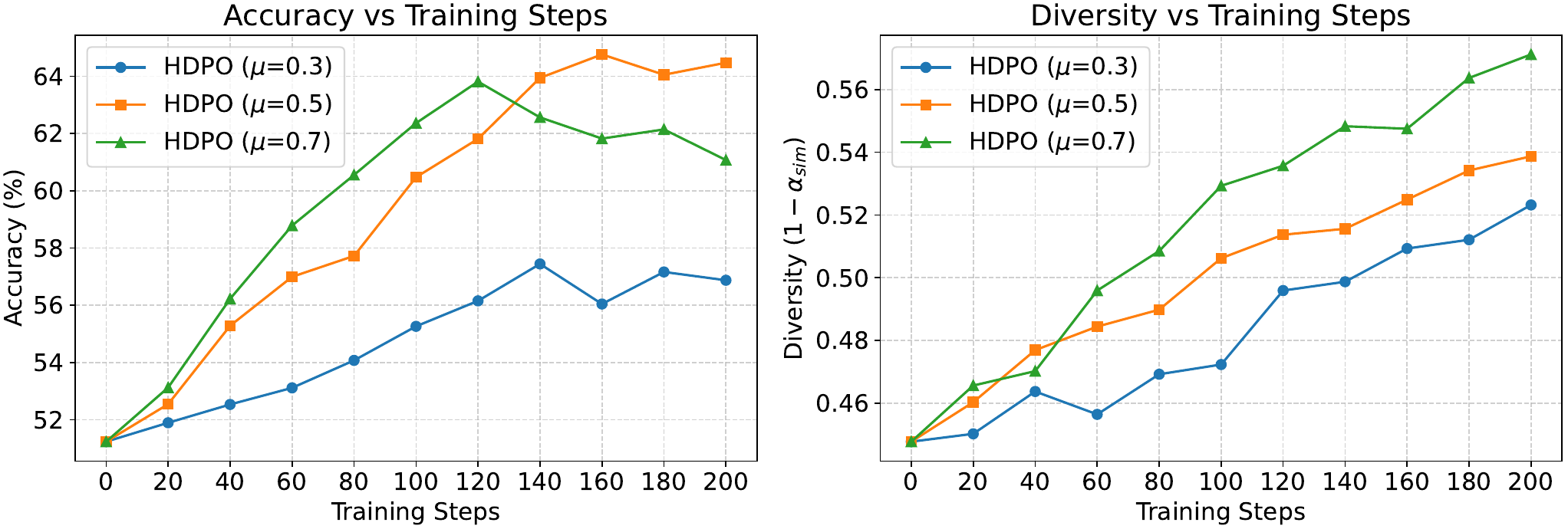}
 \caption{The effect of different weights of diversity rewards on model performance for Olympiad-Bench.}
 \label{fig:mu}
\end{center}
\vspace{-6pt}
\end{figure}

\begin{table}[t!]
\centering
\resizebox{\linewidth}{!}{
\begin{tabular}{lcccc}
\toprule
    \textbf{Variant} &  \textbf{AIME 25} & \textbf{Math} & \textbf{Olympiad} & \textbf{GPQA} \\
\midrule
GRPO (Qwen2.5-Math-7B-Instruct)    &  25.72	& 91.36	& 59.34	& 64.51   \\
HDPO (Qwen2.5-Math-7B-Instruct)    &  33.14	& 94.17	& 67.18	& 72.45   \\
\bottomrule
\end{tabular}}
\caption{Performance comparison of GRPO and HDPO with Qwen2.5-Math-7B-Instruct.}
\label{tab:base_vs_ins}
\end{table}

\begin{table*}[t]
\centering
\resizebox{\textwidth}{!}{
\begin{tabular}{lcccccccc}
\toprule
    \textbf{Methods} &   \textbf{AMC 23} & \textbf{AIME 24} & \textbf{AIME 25} & \textbf{Math} & {\textbf{Minerva}} & \textbf{Olympiad} & {\textbf{GPQA}} & \textbf{Avg.} \\
\midrule
Dr.GRPO  & 81.80 & 42.53 & 23.16 & 88.72 & 44.24 & 52.07 & 59.74  & 56.04   \\
HDPO-Dr.GRPO  & 84.75 & 52.83 & 31.55 & 93.14 & 55.55 & 58.94 & 67.67  & 63.49   \\
\hdashline[5pt/3pt]
DAPO  & 80.17 & 41.46 & 26.54 & 89.81 & 46.85 & 54.02 & 57.75  &  56.66  \\
HDPO-DAPO  & 87.22 & 50.42 & 32.25 & 94.33 & 54.61 & 59.98 & 68.73  &  63.93    \\
\hdashline[5pt/3pt]
GSPO  & 79.09 & 42.42 & 26.31 & 91.08 & 43.63 & 52.57 & 60.94  &  56.58  \\
HDPO-GSPO  & 85.31 & 51.53 & 37.32 & 93.09 & 53.62 & 56.52 & 68.14 &  63.65  \\
\bottomrule
\end{tabular}}
\caption{Experimental results using Dr.GRPO, DAPO, and GSPO as baseline algorithms.}
\label{tab:grpo_variant}
\end{table*}

\section{Trade-off between Diversity and Accuracy}
\label{appendix:balance_div_rel}
To prevent reward hacking caused by excessively low or high diversity, we control the diversity level through the hyperparameter $\mu$. We visualize the changes in accuracy and diversity during model training at different levels of diversity by setting different values of $\mu$, as shown in Figure~\ref{fig:mu}. We can observe that both lower ($\mu = 0.3$) and higher ($\mu = 0.7$) levels of diversity cause instability in model training, while the best results are achieved at $\mu = 0.5$. When the diversity level is low, the model struggles to discern distinctions among candidate solutions, potentially generating homogeneous solutions. Conversely, excessive diversity induces an overemphasis on solution heterogeneity, causing the optimization process to prioritize the diversity reward at the expense of solution correctness and thereby precipitating reward hacking behavior.

\begin{table}[t!]
\centering
\resizebox{\linewidth}{!}{
\begin{tabular}{lcccc}
\toprule
    \textbf{Variant} &  \textbf{AIME 25} & \textbf{Math} & \textbf{Olympiad} & \textbf{GPQA} \\
\midrule
HDPO ($\mathcal{M}=2$)    &  27.31 & 91.39 & 58.95 & 65.43  \\
HDPO ($\mathcal{M}=3$)    & 28.46 & 92.85 & 60.26 & 68.26   \\
HDPO ($\mathcal{M}=4$)    &  30.18 & 93.72 & 62.48 & 69.31  \\
HDPO ($\mathcal{M}=5$)    &  \textbf{31.69} & 93.28 & \textbf{64.76} & \textbf{69.89}   \\
HDPO ($\mathcal{M}=6$)    &  30.84 & \textbf{93.85} & 64.19 & 68.73   \\
HDPO ($\mathcal{M}=7$)    &  29.36 & 92.61 & 62.74 & 68.16   \\
\bottomrule
\end{tabular}}
\caption{The impact of different maximum candidate numbers $\mathcal{M}$ on model performance.}
\label{tab:maximum_number_of_candidates}
\end{table}

\section{Base Model vs. Instruction-tuned Model}
\label{appendix:base_vs_ins}
In accordance with the guidelines provided by the official Qwen2.5-Math repository, the initial analysis employed the base model Qwen2.5-Math-7B, which serves as a better starting point for fine-tuning. To investigate the generalizability of HDPO to instruction-tuned architectures, we further utilized Qwen2.5-Math-7B-Instruct as the backbone model. As demonstrated in Table~\ref{tab:base_vs_ins}, training with HDPO yields significant improvements in the reasoning performance of the instruction-tuned variant.

\section{Applying to Alternative Policy Optimization Algorithms}
\label{appendix:adapt_policy_alg}
Since HDPO is agnostic to the specific policy optimization algorithm, it can be readily adapted to other algorithms. To demonstrate this flexibility, we selected Dr.GRPO~\cite{Dr.GRPO_Oat}, DAPO~\cite{Dapo}, and GSPO~\cite{GSPO} as representative baseline algorithms for evaluation. As shown in Table~\ref{tab:grpo_variant}, integrating HDPO into each baseline algorithm yielded consistent performance improvements, thereby establishing the generalizability of our approach across diverse policy optimization frameworks.

\section{Impact of Maximum Number of Candidates}
\label{appendix:maximum_number_of_candidates}
To enhance the diversity of solutions explored by the model, we allow it to generate up to $\mathcal{M} = 5$ candidate solutions. We systematically investigate the influence of varying the maximum number of candidates $\mathcal{M}$ on model performance, with results presented in Table~\ref{tab:maximum_number_of_candidates}. The empirical analysis reveals that both insufficient and excessive values of $\mathcal{M}$ adversely affect performance. When $\mathcal{M}$ is too small, the model's exploration of the solution space remains constrained, increasing the likelihood of missing valid solutions. Conversely, when $\mathcal{M}$ is overly large, the inclusion of redundant or low-quality candidates introduces noise that can hinder the model's ability to identify the correct answer. An optimal trade-off between coverage and conciseness is achieved at $\mathcal{M} = 5$, which yields the best overall performance.

\begin{table}[t!]
\centering
\vspace{-0.1in}
\resizebox{\linewidth}{!}{
\begin{tabular}{lcccc}
\toprule
    \textbf{Variant} &  \textbf{AIME 25} & \textbf{Math} & \textbf{Olympiad} & \textbf{GPQA} \\
\midrule
EDGE-GRPO  & 19.37 (0.30) & 84.10 (0.07) & 48.79 (0.08) & 46.76 (0.14) \\
Oat  & 15.83 (0.39) & 85.34 (0.07) & 50.82 (0.10) & 44.43 (0.13) \\
SimpleRL  & 16.34 (0.58) & 85.22 (0.09) & 49.84 (0.12) & 47.31 (0.13) \\
LUFFY  & 26.61 (0.72) & 87.99 (0.14) & 56.09 (0.18) & 61.51 (0.24) \\
Critique-GRPO  & 28.44 (0.88) & 91.22 (0.19) & 61.46 (0.26) & 67.28 (0.33) \\
GRPO  & 24.55 (0.64) & 89.72 (0.12) & 53.68 (0.18) & 60.18 (0.17) \\
HDPO  & 31.69 (0.82) & 93.28 (0.16) & 64.76 (0.22) & 69.89 (0.28) \\
\bottomrule
\end{tabular}}
\caption{Analysis of the trade-off between performance and efficiency.}
\label{tab:infer_latency}
\vspace{-6pt}
\end{table}

\section{Inference Latency Analysis}
\label{appendix:infer_latency}
To further analyze the trade-off between computational resource consumption and performance, we took Qwen2.5-Math-7B as an example to analyze the inference latency. As shown in Table~\ref{tab:infer_latency}, where the performance of the model is outside the parentheses and the average inference latency (in seconds) per sample is inside the parentheses. It can be observed that HDPO increases the inference latency by less than 0.2s compared to previous SOTA methods, while achieving significant performance improvements. Additionally, the inference delay of HDPO is within 1 second, making it deployable in real-world scenarios. While the computational overhead of HDPO is a legitimate concern, it represents a principled trade-off between computational cost and reasoning quality.

\section{Comparison with Other Advanced Reasoning Strategies}
\label{appendix:comparison_other_paradigms}

To further validate the effectiveness of HDPO, we compared it against Self-Consistency (SC)~\cite{Self-Consistency} and ToTRL~\cite{ToTRL}, as detailed in Table~\ref{tab:sc_totrl_comparison}. For the SC baseline, we generated 40 reasoning paths per instance and employed majority voting to determine the final answer. The empirical results indicate that SC underperforms relative to both ToTRL and HDPO. This limitation likely stems from inadequate exploration during the inference phase, which renders the method susceptible to convergence on erroneous reasoning traces. Furthermore, HDPO demonstrates significantly superior performance compared to ToTRL. While ToTRL prioritizes answer correctness, it lacks feedback regarding the solution process. This absence of process-level supervision constrains the model's ability to explore optimal solution paths.

\begin{table}[t]
\centering
\resizebox{\linewidth}{!}{
\begin{tabular}{lrrrr}
\toprule
\textbf{Method} & \textbf{AIME 25} & \textbf{Math} & \textbf{Olympiad} & \textbf{GPQA} \\
\midrule
SC      & 19.73 & 85.34 & 48.56 & 54.12 \\
ToTRL   & 26.54 & 90.36 & 57.82 & 63.24 \\
HDPO    & 31.69 & 93.28 & 64.76 & 69.89 \\
\bottomrule
\end{tabular}
}
\caption{Performance comparison of SC, ToTRL, and HDPO with Qwen2.5-Math-7B.}
\label{tab:sc_totrl_comparison}
\end{table}

\section{Positioning of HDPO}
\label{appendix:positioning_of_hdpo}
\textbf{Policy Internalization vs. Inference-Time Search.} Unlike Tree-of-Thought (ToT)~\cite{Tree_of_Thoughts} and Self-Consistency~\cite{Self-Consistency} paradigms that rely on repeated model calls, explicit tree expansion, or post-hoc voting during inference, HDPO internalizes the ``explore–evaluate–select'' cognitive cycle directly into the policy network. This design preserves the fault-tolerance benefits of multi-path reasoning while eliminating the linear-to-exponential latency overhead inherent to search-based methods.

\noindent\textbf{Explicit Optimization for Diversity and Reliability.} Departing from accuracy-centric frameworks such as LUFFY~\cite{LUFFY} and Critique-GRPO~\cite{Critique-grpo}, which primarily optimize for outcome-level correctness or rely on external process critiques, HDPO directly addresses solution homogenization and unreliable trajectory selection. Our pipeline explicitly internalizes diversity and reliability into the policy: (1) \textit{Cold Start for Structured Reasoning} establishes the ``propose-select-think'' trajectory via SFT, rigorously filtering data by both answer correctness and \textit{candidate reliability} (validated through lightweight accuracy sampling). (2) \textit{Hint-Guided Diversified RL} injects dual optimization signals absent in prior frameworks: a sinusoidally scheduled \textit{diversity reward} penalizes semantic redundancy to prevent homogeneous convergence, while a confidence-based \textit{reliability reward} (proxied by token entropy) trains the model to autonomously rank and select the most robust candidate. By embedding these objectives directly into training, HDPO transcends correctness-centric paradigms, yielding a policy that inherently maintains broad solution coverage and fault-tolerant selection in a single forward pass.

\noindent\textbf{Synergy between Diversity and Reliability.} The diversity reward ($r_{div}$) and reliability reward ($r_{rel}$) operate on \textit{decoupled} reasoning phases: $r_{div}$ encourages variation during candidate \textit{proposal}, while $r_{rel}$ guides the \textit{selection} step. Crucially, $r_{div}$ is gated by outcome correctness, preventing exploration from compromising accuracy. The diversity scheduling strategy further ensures that diversity optimization only activates after the warm-up phase, allowing the model to first master structural compliance and baseline correctness.

\noindent\textbf{Insights on Entropy as a Reliability Proxy.} While the selection step initially relies on entropy-based confidence to choose a hint for further reasoning, the ultimate optimization signal is the Verifiable Reward (outcome correctness). If the model selects a hint with low entropy (high confidence) that leads to an incorrect solution, the RLVR mechanism assigns a negative reward. Over time, the policy learns to associate specific reasoning patterns with correctness, effectively calibrating the selection policy through trial and error, even if the raw token entropy remains uncalibrated.

Moreover, by incentivizing the generation of multiple candidate outlines before selection, the model reduces the risk of a single overconfident error dominating the process. Even if the entropy metric misranks the candidates, the correct reasoning path is more likely to be present in the candidate set compared to standard single-pass reasoning. The RL objective then reinforces the selection of hints that lead to verifiable success, gradually aligning the reliability signal with actual performance rather than raw entropy.

\section{Case Study}
\label{sec:case_study}
To further analyze the advantages of HDPO, we provide three case studies. 

\noindent \textbf{Case 1.} The first case, illustrated in Figures~\ref{fig:case1_GRPO} and~\ref{fig:case1_HDPO}, highlights a critical limitation of the traditional reasoning approach. As depicted in Figure~\ref{fig:case1_GRPO}, the traditional reasoning process verifies the constraint only at three discrete points within the interval $[0,1]$, despite the requirement that the inequality hold for all $x \in [0,1]$. Consequently, the model persists along this flawed line of reasoning and ultimately arrives at an incorrect solution. In contrast, HDPO's ``propose-select-think'' reasoning framework enables it to generate candidate solutions that incorporate Chebyshev polynomials, an essential insight that leverages the extremal properties of these polynomials to correctly determine the maximum value. 

\noindent \textbf{Case 2.} Figures~\ref{fig:case2_GRPO} and~\ref{fig:case2_HDPO} present a second illustrative example that demonstrates a different failure mode of the conventional reasoning process. In Figure~\ref{fig:case2_GRPO}, the model eventually adopts the correct solution in its third step by situating the circle within a coordinate system. However, it has already introduced an error in the second step by erroneously assuming that the distance from point $O$ to chord $\overline{AB}$ equals 4. This error stems from the model's failure to recognize the proper solution strategy at the outset. By contrast, the HDPO method depicted in Figure~\ref{fig:case2_HDPO} introduces coordinate geometry as a candidate approach from the beginning and maintains this coherent line of reasoning throughout, ultimately yielding the correct solution.

\noindent \textbf{Case 3 (Failure Case).} In addition, we also present a failure case of HDPO, as shown in Figure~\ref{fig:case3_failure_case}. In this example, the model chose the first candidate solution, but this solution did not take into account the use of complex number theory, which represents the fundamental approach to solving this problem. Consequently, the model made an erroneous assumption during reasoning. Specifically, it incorrectly concluded that angle pairs such as $(8^\circ, 352^\circ)$ and $(16^\circ, 344^\circ)$ possess opposite cosine values, failing to incorporate the complex number perspective into its analysis. The case shows that when the initial set of candidate solutions is generated without the correct answer, the subsequent reasoning is compromised. Therefore, steering the model toward the generation of more pertinent candidate solutions could enhance the reasoning capabilities of LLMs.

\begin{figure*}[t]
\begin{center}
    \begin{tcolorbox}[
      title=Structured Reasoning Prompt,
      colback=white,
      colframe=black!70,
      fontupper=\small\ttfamily,
      coltitle=white,
      colbacktitle=blue!50!black,
      fonttitle=\ttfamily,
      boxrule=0.8pt,
    ]
    Given a problem, please first consider all possible candidate solutions, and place an overview of all candidate solutions within <Candidate Solutions></Candidate Solutions>. Different solution overviews should be separated by ``\textbackslash n\textbackslash n''.

    \vspace{12pt}
    
    1. The overview of the solution should be as concise as possible, containing 1 to 5 sentences. It should only elaborate on the high-level strategies and concepts, without going into specific calculations.

    \vspace{12pt}
    
    2. Different candidate solutions should exist differences at the method level, so do not output similar solution overviews.

    \vspace{12pt}
    
    3. Do not use line breaks within the same candidate solution overview, and use ``\textbackslash n\textbackslash n'' to separate different candidate solutions.

    \vspace{12pt}
    
    4. You can propose 1 to 5 candidate solutions, each preceded by the serial number enclosed in square brackets, such as ``[1]'', ``[2]'', ``[3]'', etc. Each candidate solution overview is a core summary of the proposed solution.

    \vspace{12pt}
    
    After thinking through the candidate solution overviews, choose the most reliable one and place the step number of the chosen solution (e.g., ``[1]'') within <selected></selected>. Next, proceed with step-by-step reasoning based on the chosen solution. Place the reasoning process in the <thinking></thinking> and put your final answer in \textbackslash \textbackslash boxed\{\}.

    \vspace{12pt}
    
    Problem: \{\}
    \end{tcolorbox}
\end{center}
\vspace{-6pt}
\caption{The prompt used in structured reasoning.}
\label{fig:prompt_reasoning}
\end{figure*}

\begin{figure*}[t]
\begin{center}
    \begin{tcolorbox}[
      title=Relevance Judgment Prompt,
      colback=white,
      colframe=black!70,
      fontupper=\small\ttfamily,
      coltitle=white,
      colbacktitle=blue!50!black,
      fonttitle=\ttfamily,
      boxrule=0.8pt,
    ]
    Given a question and its corresponding two candidate solution outlines, please determine whether these two candidate solutions are the same. We consider them to be the same when they meet the following two conditions.

    \vspace{12pt}
    
    1. They use the same branch of mathematics or specific theorem.
    
    2. They rely on the same underlying logical dependencies, although the expressions may differ.

    \vspace{12pt}
    
    If the two candidate solutions are the same, output "Yes", otherwise output "No", do not output redundant content.

    \vspace{12pt}
    
    Question: \{question\}

    Candidate Solution 1: \{candidate solution 1\}

    Candidate Solution 2: \{candidate solution 2\}
    \end{tcolorbox}
\end{center}
\vspace{-6pt}
\caption{The prompt used for solution relevance judgment.}
\label{fig:LLM_Div}
\end{figure*}

\begin{figure*}[htbp]
\begin{center}
\begin{tcolorbox}[
  title=Case 1 (Generation Model: Qwen2.5-Math-7B-GRPO),
  colback=white,
  colframe=black!70,
  fontupper=\small\ttfamily,
  coltitle=white,
  colbacktitle=red,
  fonttitle=\ttfamily,
  boxrule=0.8pt,
]
\textbf{Question:} \\
Let $a,$ $b,$ $c$ be real numbers such that
\[|ax^2 + bx + c| \le 1\]for all $0 \le x \le 1.$  Find the largest possible value of $|a| + |b| + |c|.$ \\
\vspace{-5pt} \hrule \vspace{5pt}
\textbf{Response:} \\
<thinking>

To solve the problem, we need to find the maximum value of \(|a| + |b| + |c|\) given the constraint \(|ax^2 + bx + c| \le 1\) for all \(0 \le x \le 1\).

We can start by evaluating the polynomial at specific points within the interval \([0, 1]\). The points \(x = 0\), \(x = \frac{1}{2}\), and \(x = 1\) are good choices because they will give us a system of equations that we can use to determine \(a\), \(b\), and \(c\).

1. At \(x = 0\), we have \(|c| \le 1\).

2. At \(x = 1\), we have \(|a + b + c| \le 1\).

3. At \(x = \frac{1}{2}\), we have \(| \frac{a}{4} + \frac{b}{2} + c | \le 1\).

We need to find the values of \(a\), \(b\), and \(c\) that maximize \(|a| + |b| + |c|\) while satisfying these constraints. Let's consider the possible values for \(c\). Since \(|c| \le 1\), we can set \(c = 1\) or \(c = -1\) to maximize the sum.

Let's start with \(c = 1\). Then we have:

1. \(|a + b + 1| \le 1\), which implies \(-2 \le a + b \le 0\).

2. \(| \frac{a}{4} + \frac{b}{2} + 1 | \le 1\), which implies \(-2 \le \frac{a}{4} + \frac{b}{2} + 1 \le 0\), or \(-3 \le \frac{a}{4} + \frac{b}{2} \le -1\).

We can solve these inequalities to find the values of \(a\) and \(b\). Let's consider the boundary cases to maximize \(|a| + |b| + |c|\).

If we set \(a = 4\) and \(b = -6\), we get:
1. \(|4 - 6 + 1| = |-1| = 1 \le 1\), which satisfies the constraint.
2. \(| \frac{4}{4} + \frac{-6}{2} + 1 | = |1 - 3 + 1| = |-1| = 1 \le 1\), which satisfies the constraint.

So, \(|a| + |b| + |c| = |4| + |-6| + |1| = 4 + 6 + 1 = 11\).

Let's check if there are any other possible values that could give a larger sum. If we set \(c = -1\), we get:

1. \(|a + b - 1| \le 1\), which implies \(0 \le a + b \le 2\).

2. \(| \frac{a}{4} + \frac{b}{2} - 1 | \le 1\), which implies \(0 \le \frac{a}{4} + \frac{b}{2} \le 2\), or \(0 \le \frac{a}{4} + \frac{b}{2} \le 2\).

If we set \(a = -4\) and \(b = 6\), we get:

1. \(|-4 + 6 - 1| = |1| = 1 \le 1\), which satisfies the constraint.

2. \(| \frac{-4}{4} + \frac{6}{2} - 1 | = |-1 + 3 - 1| = |1| = 1 \le 1\), which satisfies the constraint.

So, \(|a| + |b| + |c| = |-4| + |6| + |-1| = 4 + 6 + 1 = 11\).

Therefore, the largest possible value of \(|a| + |b| + |c|\) is \(\boxed{11}\).

</thinking>

\boxed{11}  \textcolor{red}{\ding{56}} \textcolor{darkgreen}{\textbf{Correct Answer:} \bm{$17$}}
\end{tcolorbox}
\end{center}
\vspace{-6pt}
\caption{The output produced by Qwen2.5-Math-7B following GRPO training with standard chain-of-thought reasoning in the first test case.}
\label{fig:case1_GRPO}
\end{figure*}

\begin{figure*}[t]
\begin{center}
\begin{tcolorbox}[
  title=Case 1 (Generation Model: Qwen2.5-Math-7B-HDPO),
  colback=white,
  colframe=black!70,
  fontupper=\small\ttfamily,
  coltitle=white,
  colbacktitle=green!70!black,
  fonttitle=\ttfamily,
  boxrule=0.8pt,
]
\textbf{Question:} \\
Let $a,$ $b,$ $c$ be real numbers such that
\[|ax^2 + bx + c| \le 1\]for all $0 \le x \le 1.$  Find the largest possible value of $|a| + |b| + |c|.$ \\
\vspace{-5pt} \hrule \vspace{5pt}
\textbf{Response:} \\
<Candidate Solutions>

[1] Use the fact that the quadratic polynomial is bounded by 1 in absolute value on [0,1], and evaluate it at key points such as x=0, x=1, and x=0.5 to derive constraints on a, b, c. Then express a, b, c in terms of these values and use linear programming or symmetry arguments to maximize |a| + |b| + |c|.

[2] Consider the polynomial as a function in a finite-dimensional space and use the extreme values of the polynomial on [0,1] to form a system of inequalities. Apply the method of Lagrange multipliers to maximize the linear functional |a| + |b| + |c| subject to the quadratic constraint.

[3] Assume symmetry or specific forms (e.g., a=0, or b=0) to reduce complexity and test boundary cases. However, this may miss the global maximum if the optimal polynomial is not symmetric or has non-zero a and b.

[4] Use Chebyshev polynomials, which are known to be extremal with respect to maximum deviation from zero, and map the standard Chebyshev polynomial on [-1,1] to the interval [0,1]. Scale and shift the Chebyshev polynomial to satisfy the bound $|P(x)| \leq 1$ and compute the coefficients to find the maximum of |a| + |b| + |c|.

[5] Use polynomial interpolation and optimization: since the polynomial is bounded on [0,1], consider the values at three points (e.g., 0, 1/2, 1) and use the resulting system to express coefficients, then apply bounds and optimization techniques to maximize the sum of absolute values.

</Candidate Solutions>

<selected>[4]</selected>

<thinking>

We are given that $ |ax^2 + bx + c| \le 1 $ for all $ 0 \le x \le 1 $, and we are to find the maximum value of $ |a| + |b| + |c| $.

The key insight is to use the properties of Chebyshev polynomials, which are known for being extremal with respect to the maximum norm. The Chebyshev polynomial of degree 2 on $[-1,1]$ is $ T_2(x) = 2x^2 - 1 $, which oscillates between -1 and 1 at three points.

We want to map this polynomial to the interval $[0,1]$. Let $ t = 2x - 1 $, which maps $ x \in [0,1] $ to $ t \in [-1,1] $. Then $ x = \frac{t+1}{2} $, and we can substitute into $ T_2(t) = 2t^2 - 1 $:

\[
P(x) = T_2(2x - 1) = 2(2x - 1)^2 - 1 = 2(4x^2 - 4x + 1) - 1 = 8x^2 - 8x + 2 - 1 = 8x^2 - 8x + 1.
\]

So $ a = 8 $, $ b = -8 $, $ c = 1 $, and $ |a| + |b| + |c| = 8 + 8 + 1 = 17 $.

Now, we check that $ |P(x)| \le 1 $ for $ 0 \le x \le 1 $. Since $ P(x) = T_2(2x - 1) $, and $ T_2(t) $ has maximum absolute value 1 on $[-1,1]$, it follows that $ |P(x)| \le 1 $ for all $ x \in [0,1] $.

Is this the maximum possible? Suppose there exists another polynomial $ Q(x) = ax^2 + bx + c $ with $ |Q(x)| \le 1 $ on $[0,1]$ and $ |a| + |b| + |c| > 17 $. By the properties of Chebyshev polynomials, the maximum of $ |a| + |b| + |c| $ under the given constraint is achieved at the extremal polynomial, which is a scaled and shifted Chebyshev polynomial. Since $ P(x) = 8x^2 - 8x + 1 $ achieves $ |a| + |b| + |c| = 17 $ and satisfies the bound, it must be the maximum.

Thus, the largest possible value is $ \boxed{17} $.

</thinking>

\boxed{17} \textcolor{darkgreen}{\ding{51}}

\end{tcolorbox}
\end{center}
\vspace{-6pt}
\caption{The output produced by Qwen2.5-Math-7B following HDPO training using ``propose-select-think'' structured reasoning in the first test case.}
\label{fig:case1_HDPO}
\end{figure*}

\begin{figure*}[htbp]
\begin{center}
\begin{tcolorbox}[
  title=Case 2 (Generation Model: Qwen2.5-Math-7B-GRPO),
  colback=white,
  colframe=black!70,
  fontupper=\small\ttfamily,
  coltitle=white,
  colbacktitle=red,
  fonttitle=\ttfamily,
  boxrule=0.8pt,
]
\textbf{Question:} \\
Circle $O$ has radius 6. Point $P$ lies outside circle $O$, and the shortest distance from $P$ to circle $O$ is 4. Chord $\overline{A B}$ is parallel to $\overleftrightarrow{O P}$, and the distance between $\overline{A B}$ and $\overleftrightarrow{O P}$ is 2 . Compute $P A^{2}+P B^{2}$. \\
\vspace{-5pt} \hrule \vspace{5pt}
\textbf{Response:} \\
<thinking>
To solve the problem, we need to find the value of \(PA^2 + PB^2\). Here's the step-by-step reasoning:

1. **Understand the Geometry:**
   - Circle \(O\) has a radius of 6.
   
   - Point \(P\) is 4 units away from the circle, so the distance from \(P\) to the center \(O\) is \(6 + 4 = 10\).
   
   - Chord \(\overline{AB}\) is parallel to \(\overleftrightarrow{OP}\) and the distance between \(\overline{AB}\) and \(\overleftrightarrow{OP}\) is 2.

2. **Find the Length of Chord \(\overline{AB}\):**

   - Since \(\overline{AB}\) is parallel to \(\overleftrightarrow{OP}\) and the distance between them is 2, the perpendicular distance from \(O\) to \(\overline{AB}\) is \(6 - 2 = 4\) or \(6 + 2 = 8\). But since \(P\) is 4 units away from the circle and \(\overline{AB}\) is parallel to \(\overleftrightarrow{OP}\), the perpendicular distance from \(O\) to \(\overline{AB}\) must be 4 (as 8 would place \(\overline{AB}\) outside the circle).
   
   - Using the Pythagorean theorem in the right triangle formed by the radius, the perpendicular distance, and half the chord, we get:
     \[
     \left(\frac{AB}{2}\right)^2 = 6^2 - 4^2 = 36 - 16 = 20 \implies \frac{AB}{2} = \sqrt{20} = 2\sqrt{5} \implies AB = 4\sqrt{5}
     \]

3. **Find the Coordinates:**

   - Place the circle centered at \(O(0, 0)\) and point \(P\) at \((10, 0)\).
   
   - The chord \(\overline{AB}\) is parallel to the x-axis and 2 units below the x-axis (since the distance from \(O\) to \(\overline{AB}\) is 4).
   
   - So, the coordinates of \(A\) and \(B\) can be \((-2\sqrt{5}, -2)\) and \((2\sqrt{5}, -2)\) respectively.

4. **Calculate \(PA^2\) and \(PB^2\):**

   - Using the distance formula, we get:
     \[
     PA^2 = (10 - (-2\sqrt{5}))^2 + (0 - (-2))^2 = (10 + 2\sqrt{5})^2 + 2^2 = 100 + 40\sqrt{5} + 20 + 4 = 124 + 40\sqrt{5}
     \]
     \[
     PB^2 = (10 - 2\sqrt{5})^2 + (0 - (-2))^2 = (10 - 2\sqrt{5})^2 + 2^2 = 100 - 40\sqrt{5} + 20 + 4 = 124 - 40\sqrt{5}
     \]
     
   - Adding \(PA^2\) and \(PB^2\):
     \[
     PA^2 + PB^2 = (124 + 40\sqrt{5}) + (124 - 40\sqrt{5}) = 248
     \]

So, the final answer is:
\[
\boxed{248}
\]
</thinking>

\boxed{248} \textcolor{red}{\ding{56}} \textcolor{darkgreen}{\textbf{Correct Answer:} \bm{$272$}}
\end{tcolorbox}
\end{center}
\vspace{-6pt}
\caption{The output produced by Qwen2.5-Math-7B following GRPO training with standard chain-of-thought reasoning in the second test case.}
\label{fig:case2_GRPO}
\end{figure*}

\begin{figure*}[t]
\begin{center}
\begin{tcolorbox}[
  title=Case 2 (Generation Model: Qwen2.5-Math-7B-HDPO),
  colback=white,
  colframe=black!70,
  fontupper=\small\ttfamily,
  coltitle=white,
  colbacktitle=green!70!black,
  fonttitle=\ttfamily,
  boxrule=0.8pt,
]
\textbf{Question:} \\
Circle $O$ has radius 6. Point $P$ lies outside circle $O$, and the shortest distance from $P$ to circle $O$ is 4. Chord $\overline{A B}$ is parallel to $\overleftrightarrow{O P}$, and the distance between $\overline{A B}$ and $\overleftrightarrow{O P}$ is 2 . Compute $P A^{2}+P B^{2}$. \\
\vspace{-5pt} \hrule \vspace{5pt}
\textbf{Response:} \\
<Candidate Solutions>

[1] Use geometric symmetry and properties of circles and chords. Since $AB$ is parallel to $OP$ and the distance between them is 2, the perpendicular distance from $O$ to $AB$ is either $4 + 2 = 6$ or $4 - 2 = 2$. This approach relies on parameterizing the circle and solving systems of equations.

[2] Use vector geometry: represent points $A$ and $B$ on the circle with $AB$ parallel to vector $\vec{OP}$, and the perpendicular distance between the lines is 2.

[3] Use coordinate geometry by placing the center of circle $O$ at the origin and point $P$ along the x-axis. Since chord $AB$ is parallel to $\overleftrightarrow{OP}$, it is horizontal, and the distance between $AB$ and $\overleftrightarrow{OP}$ is 2, so $AB$ is either at $y = 2$ or $y = -2$.

</Candidate Solutions>

<selected>[3]</selected>

<thinking>

We place the center of circle $O$ at the origin $(0, 0)$. Since the radius is 6, the circle has equation $x^2 + y^2 = 36$.

The shortest distance from point $P$ to the circle is 4. Since the radius is 6, this means that the distance from $O$ to $P$ is $6 + 4 = 10$. We can place $P$ at $(10, 0)$, so that it lies on the x-axis.

The line $\overleftrightarrow{OP}$ is the x-axis (since $O = (0,0)$, $P = (10,0)$). The chord $AB$ is parallel to $\overleftrightarrow{OP}$, so it is horizontal. The distance between $AB$ and $\overleftrightarrow{OP}$ is 2. Since $AB$ could be above or below the x-axis, we consider $y = 2$ and $y = -2$. By symmetry, the value of $PA^2 + PB^2$ will be the same in both cases, so we can choose $y = 2$.

Now, find the points $A$ and $B$ where $y = 2$ intersects the circle $x^2 + y^2 = 36$:
$$
x^2 + 2^2 = 36 \Rightarrow x^2 + 4 = 36 \Rightarrow x^2 = 32 \Rightarrow x = \pm \sqrt{32} = \pm 4\sqrt{2}
$$
So, $A = (4\sqrt{2}, 2)$, $B = (-4\sqrt{2}, 2)$

Now compute $PA^2$ and $PB^2$, where $P = (10, 0)$

First, $PA^2$:
$$
PA^2 = (10 - 4\sqrt{2})^2 + (0 - 2)^2 = (10 - 4\sqrt{2})^2 + 4
$$
Expand:
$$
(10 - 4\sqrt{2})^2 = 100 - 80\sqrt{2} + 16 \cdot 2 = 100 - 80\sqrt{2} + 32 = 132 - 80\sqrt{2}
$$
So:
$$
PA^2 = 132 - 80\sqrt{2} + 4 = 136 - 80\sqrt{2}
$$

Now $PB^2$:
$$
PB^2 = (10 + 4\sqrt{2})^2 + (0 - 2)^2 = (10 + 4\sqrt{2})^2 + 4
$$
Expand:
$$
(10 + 4\sqrt{2})^2 = 100 + 80\sqrt{2} + 32 = 132 + 80\sqrt{2}
$$
So:
$$
PB^2 = 132 + 80\sqrt{2} + 4 = 136 + 80\sqrt{2}
$$

Now add $PA^2 + PB^2$:
$$
(136 - 80\sqrt{2}) + (136 + 80\sqrt{2}) = 136 + 136 = 272
$$

The $\sqrt{2}$ terms cancel out.

Thus, the answer is $ \boxed{272} $

</thinking>

\boxed{272} \textcolor{darkgreen}{\ding{51}}

\end{tcolorbox}
\end{center}
\vspace{-6pt}
\caption{The output produced by Qwen2.5-Math-7B following HDPO training using ``propose-select-think'' structured reasoning in the second test case.}
\label{fig:case2_HDPO}
\end{figure*}

\begin{figure*}[t]
\begin{center}
\begin{tcolorbox}[
  title=Case 3 (Generation Model: Qwen2.5-Math-7B-HDPO),
  colback=white,
  colframe=black!70,
  fontupper=\small\ttfamily,
  coltitle=white,
  colbacktitle=red,
  fonttitle=\ttfamily,
  boxrule=0.8pt,
]
\textbf{Question:} \\
Compute $\sin ^{2} 4^{\circ}+\sin ^{2} 8^{\circ}+\sin ^{2} 12^{\circ}+\cdots+\sin ^{2} 176^{\circ}$. \\
\vspace{-5pt} \hrule \vspace{5pt}
\textbf{Response:} \\
<Candidate Solutions>

[1] Use the identity $\sin^2 \theta = \frac{1 - \cos 2\theta}{2}$ to rewrite each term in the sum. Then apply the sum of cosine series formula for angles in arithmetic sequence, simplifying the resulting expression using symmetry and periodicity of the cosine function.

[2] Interpret the sum as a discrete approximation of an integral of $\sin^2 x$ over a interval, and use numerical methods or symmetry to estimate the value. However, this approach lacks precision and relies on approximation, making it unsuitable for exact computation.

[3] Use complex exponentials to express $\sin^2 \theta$ and convert the sum into a sum of powers of roots of unity, then apply properties of Gauss sums or Fourier analysis on finite groups.

[4] Use a recursive or computational algorithm to evaluate each term numerically and sum them, but this is impractical for exact mathematical reasoning and lacks elegance and precision.

</Candidate Solutions>

<selected>[1]</selected>

<thinking>

We are given the sum:
$$
\sin^2 4^\circ + \sin^2 8^\circ + \sin^2 12^\circ + \cdots + \sin^2 176^\circ
$$
This is a sum of $\sin^2 \theta$ for $\theta = 4k^\circ$ where $k = 1, 2, \ldots, 44$, since $4 \times 44 = 176$.

We use the identity:
$$
\sin^2 \theta = \frac{1 - \cos 2\theta}{2}
$$
Applying this to each term:
$$
\sin^2 (4k^\circ) = \frac{1 - \cos (8k^\circ)}{2}
$$
So the entire sum becomes:
$$
\sum_{k=1}^{44} \sin^2 (4k^\circ) = \sum_{k=1}^{44} \frac{1 - \cos (8k^\circ)}{2} = \frac{1}{2} \sum_{k=1}^{44} 1 - \frac{1}{2} \sum_{k=1}^{44} \cos (8k^\circ)
$$
$$
= \frac{1}{2} \cdot 44 - \frac{1}{2} \sum_{k=1}^{44} \cos (8k^\circ) = 22 - \frac{1}{2} \sum_{k=1}^{44} \cos (8k^\circ)
$$
Now we need to compute:
$$
\sum_{k=1}^{44} \cos (8k^\circ)
$$
The sum of $\cos (8k^\circ)$ for $k = 1$ to $44$ can be evaluated using symmetry.

Note that the angles $8k^\circ$ for $k = 1$ to $44$ are evenly spaced from $8^\circ$ to $352^\circ$, which is half the circle (since $8 \times 44 = 352$). These angles are symmetric about $180^\circ$, and the cosine function is even but symmetric about $180^\circ$, so the sum of cosines over a symmetric arithmetic sequence about $180^\circ$ with evenly spaced steps is zero.

To see this: the sequence $8^\circ, 16^\circ, \ldots, 352^\circ$ is symmetric about $180^\circ$, and $\cos(360^\circ - x) = \cos x$, so pairs like $(8^\circ, 352^\circ)$, $(16^\circ, 344^\circ)$, etc., have equal and opposite cosines. Since there are 44 terms (even number), they can be perfectly paired, and each pair sums to zero.

Therefore:
$$
\sum_{k=1}^{44} \cos (8k^\circ) = 0
$$
So the original sum is:
$$
22 - \frac{1}{2} \cdot 0 = 22
$$
</thinking>

\boxed{22} \textcolor{red}{\ding{56}} \textcolor{darkgreen}{\textbf{Correct Answer:} \bm{$\frac{45}{2}$}}

\end{tcolorbox}
\end{center}
\vspace{-6pt}
\caption{A failure case of Qwen2.5-Math-7B following HDPO training, where the candidate solutions did not include the most reliable solution.}
\label{fig:case3_failure_case}
\end{figure*}

\end{document}